\ifdefined\pdfminorversion\pdfminorversion=7\fi
\documentclass[10pt,letterpaper]{article}
\usepackage{PRIMEarxiv}
\usepackage[utf8]{inputenc}
\usepackage[T1]{fontenc}
\usepackage{times}
\usepackage{xspace}
\usepackage[dvipsnames]{xcolor}
\usepackage{graphicx}
\usepackage{amsmath,amssymb}
\usepackage{booktabs,makecell,multirow,tabularx}
\usepackage{pifont}
\usepackage{microtype}
\usepackage[export]{adjustbox}
\usepackage[numbers,sort&compress]{natbib}
\usepackage[hyphens]{url}
\usepackage[format=plain,labelformat=simple,labelsep=period,font=small]{caption}
\usepackage[shortlabels,inline]{enumitem}
\usepackage{placeins}
\usepackage{wrapfig}
\usepackage{needspace}
\definecolor{linkblue}{rgb}{0.21,0.49,0.74}
\usepackage[breaklinks,colorlinks,allcolors=linkblue]{hyperref}
\usepackage[capitalize]{cleveref}
\crefname{section}{Sec.}{Secs.}
\Crefname{section}{Section}{Sections}
\crefname{table}{Tab.}{Tabs.}
\Crefname{table}{Table}{Tables}

\setlist[itemize]{leftmargin=*,topsep=4pt,itemsep=2pt,parsep=0pt}

\newcommand{\ours}{AgroWM\xspace}

\newcommand{\clock}{PhenoClock\xspace}
\newcommand{\script}{Growth Script\xspace}
\newcommand{\scripts}{Growth Scripts\xspace}
\newcommand{\tutor}{Hindsight Tutor\xspace}
\newcommand{\renderer}{Rooted Renderer\xspace}

\newcommand{\ZetengLiEmail}{}
\newcommand{\YinkunGuoEmail}{}
\newcommand{\ZhenongJinEmail}{}
\newcommand{\PekingUniversityDepartment}{}
\newcommand{\PekingUniversityAddress}{}

\author{%
  \normalfont
  Wenkai Ji$^{1}$, Zeteng Li$^{2}$, Yinkun Guo$^{2}$,
  Jie Zhou$^{1,}$\thanks{Corresponding authors.}\kern0.35em,
  Liang He$^{1}$, Zhenong Jin$^{2,*}$ \\[0.5em]
  \normalfont\small $^{1}$School of Computer Science and Technology, East China Normal University, Shanghai \\
  \normalfont\small $^{2}$\ifdefempty{\PekingUniversityDepartment}{}{\PekingUniversityDepartment, }Peking University\ifdefempty{\PekingUniversityAddress}{}{, \PekingUniversityAddress} \\[0.4em]
  \normalfont\small\texttt{\{jzhou, lhe\}@cs.ecnu.edu.cn}
  \ifdefempty{\ZetengLiEmail}{}{\\\normalfont\small\texttt{\ZetengLiEmail}}
  \ifdefempty{\YinkunGuoEmail}{}{\\\normalfont\small\texttt{\YinkunGuoEmail}}
  \ifdefempty{\ZhenongJinEmail}{}{\\\normalfont\small\texttt{\ZhenongJinEmail}} \\
    \textcolor{red}{Github: \url{https://github.com/ECNU-ICALK/ArgoWM}} \\
    \textcolor{red}{Model: \url{https://huggingface.co/models/ecnu-icalk/ArgoWM}} \\
}

\newif\ifcombinedanalysis
\combinedanalysistrue
\newif\ifmaize
\maizetrue

\newcommand{\papertitle}{Can World Models Predict Rice Growth? Toward Agricultural World Models}
\title{\papertitle}
\date{}
\hypersetup{
  pdftitle={Can World Models Predict Rice Growth? Toward Agricultural World Models},
  pdfauthor={Wenkai Ji, Zeteng Li, Yinkun Guo, Jie Zhou, Liang He, Zhenong Jin},
  pdfsubject={Agricultural world modeling},
  pdfkeywords={Agricultural world models, plant growth forecasting, phenological reasoning}
}

\begin{document}
\maketitle
\begin{abstract}
World models can imagine how a physical scene evolves over the next few seconds, but growers need to know how a crop will develop over weeks to months: will this rice plant have headed a month from now, and how will it look at harvest?
We introduce \emph{agricultural world modeling}, the task of predicting how the same plant will look at a requested horizon of days to months, from one photo of it, its cultivar, and its camera view.
In preliminary analysis, even state-of-the-art image generators get this wrong: they often grow the plant too little or give it the color and form of the wrong growth stage.
We argue that they do not work out the plant's developmental state, which the photo reveals only indirectly.
We therefore propose \ours, which reasons about development before it draws growth.
Its reasoning module, \clock, works out the current growth stage, predicts the future one, and writes a \script with numeric targets for height, plant area, and color; its generation module, the \renderer, then edits the photo to follow this script, adding new organs and colors to the photographed plant.
Because pixel scores can prefer an unchanged copy of the photo, we instead check each forecast's traits and developmental timing against the plant's own later photos.
On rice, 54.6\% of the forecasts of \ours are matched to the target stage (Stage R@1), against 34.8\% for GPT Image 2.5, 24.0\% for Qwen Image 2.1, and 47.4\% for the strongest domain-specific baseline, the same backbone fine-tuned without \scripts; 9 to 20 weeks ahead, it still reaches 46.4\%, against 35.7\% for that baseline.
\ifmaize
On maize, a crop with a different body plan and stage scale, the same design again scores best on all four measures.
\fi
\end{abstract}

\section{Introduction}
\label{sec:intro}

\begin{figure}[!t]
  \centering
  \includegraphics[width=\linewidth]{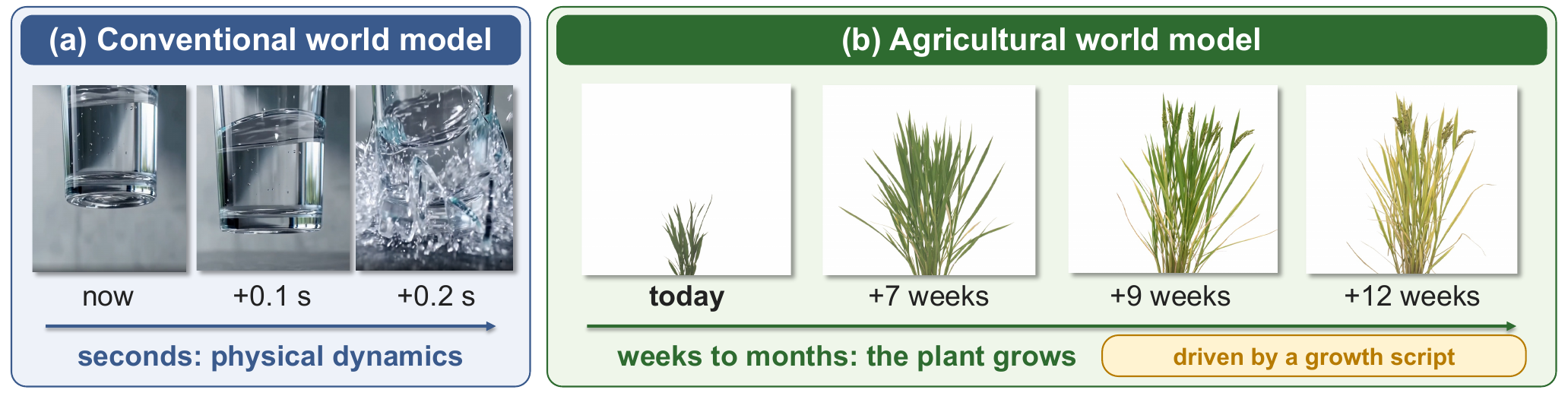}
  \caption{\textbf{From physical dynamics to growth.}
  (a)~A conventional world model predicts rapid physical events.
  (b)~From one photo, an agricultural world model predicts how the plant will look at each requested horizon of weeks to months, as it moves through growth stages.}
  \label{fig:teaser}
\end{figure}

World models~\cite{brooks2024sora,hu2023gaia,yang2024unisim} let machines imagine how a physical scene evolves over the next few seconds, as when a glass falls, strikes a surface, and shatters (\cref{fig:teaser}a).
Questions about growing food instead require forecasts over weeks to months: how tall will this plant grow, how much will its canopy spread, and when will it flower and fill its grain?
For rice, which feeds more than half of the world's population~\cite{muthayya2014overview}, the answers otherwise take up to a season to arrive.
We ask whether a world model can give them in advance: from one photo of a plant, its cultivar, and its camera view, predict how the same plant will look at a requested horizon of days to months (\cref{fig:teaser}b).
We call this task \emph{agricultural world modeling}.
Its answer is a photo, which, unlike a single number, can be checked by eye and measured for any visible trait.

The task is harder than it looks, as shown by a preliminary study of 64 rice cases: 1--4 weeks (22 cases), 5--8 weeks (24), and 9--20 weeks (18) (\cref{sec:analysis}).
\emph{First, two state-of-the-art image generators fail at it}: Qwen Image 2.1 hardly changes the plant, although real plants grow about tenfold in projected area over 9 to 20 rounded weeks (median $11.2\times$), and GPT Image 2 grows it too little at long horizons, shows senescence-related color changes at the wrong stage, and redraws it toward a generic rice plant (\cref{fig:analysis}).
\emph{Second, the horizon alone does not decide the future stage}: of the 18 cases with horizons of 9 to 20 weeks, 6 plants have not yet headed by the target date, while 9 are already filling grain or ripening.
A forecaster must therefore infer each plant's developmental state, which the photo reveals only indirectly.
\emph{Third, common image scores reward doing nothing}: a copy of today's photo has the highest SSIM in \cref{tab:main}, and Qwen Image 2.1, which barely changes the plant, the highest PSNR.

A reliable growth forecast must get the individual plant, its future traits, and their timing right.
Existing methods address only parts of this problem.
Time-lapse generators~\cite{yuan2025magictime,yuan2024chronomagic} draw plausible growth from text, with no particular plant or date to match, and keyframe interpolation~\cite{wang2025inbetweening} and 4D reconstruction~\cite{luo2026growflow} are given the later images that a forecast has to predict.
Plant-growth GANs~\cite{drees2021temporal,drees2024datadriven,wang2025maize} come closest, but they map elapsed time, treatment, or simulated biomass to pixels without reasoning about development.
Agricrafter~\cite{agricrafter} generates controllable videos of a crop's full growth cycle, but a coherent sequence alone does not show whether a given plant reaches the correct stage within the requested horizon.
Crop models~\cite{bouman2001oryza,jones2003dssat} do track development using temperature and day length, but require cultivar- and site-specific parameters and do not directly generate the future appearance of an individual plant.
Our approach follows an agronomist’s reasoning by first predicting the plant’s future developmental stage and then generating an image consistent with that prediction. For example, if a plant is currently tillering and is expected to reach jointing in four weeks, the forecast image should show elongated stems and a taller plant.

We turn this insight into \ours, an agricultural world model with two modules (\cref{fig:pipeline}).
First, its reasoning module, \clock, is a vision-language model that works out the plant's current stage, predicts its stage at the horizon, and writes this chain of thought (CoT) as a \script.
The script is a short, structured note: the current and future stage and, for each visible trait such as height, plant area, and color, today's value, the target value, and the change in words.
Since no dataset records such notes, \clock learns from scripts written in hindsight: for each training pair, the stages and values come from records and measurements of the real future, and a language model adds the words (\cref{sec:planner}).
Second, its generation module, the \renderer, then edits today's photo as the script says; it is prompted and trained to keep the plant's form, framing, and scale, on pairs chosen to reward real growth (\cref{sec:renderer}).

To judge forecasts, we check what matters to a grower: the same plant, at the right size and color, at the right stage of development (\cref{sec:protocol}).
Timing is checked against the plant's own photo series: Stage R@1 is the share of forecasts whose most similar photo of the plant lies in the target stage.
On held-out plants of the public rice dataset WGSR~\cite{tang2023wgsr}, \ours reaches 54.6\% Stage R@1, against 34.8\% for GPT Image 2.5 and 24.0\% for Qwen Image 2.1, two state-of-the-art generators, and 47.4\% for the strongest domain-specific model, the same backbone fine-tuned on the same pairs without \scripts (\cref{tab:main}).
Reasoning about development thus adds 7.2 points (95\% CI 5.2--9.2), and 10.7 points 9 to 20 weeks ahead (\cref{fig:horizon}).
\ifmaize
On maize, which grows a single stalk with a tassel instead of tillers and panicles, the same design again scores best, so the approach is not tied to one crop (\cref{sec:exp-maize}).
\fi

\noindent In summary, we make three contributions:
\begin{itemize}
  \item 
  We explore agricultural world modeling and find that the current image generators do not pace a plant's growth to the horizon, reach its target stage, or keep the individual.
  \item 
  We propose \ours, which reasons before it renders: \clock forecasts the developmental stage with the plant's height, area, and color as an agricultural CoT learned from scripts written in hindsight, and the \renderer turns this forecast into an edited photo of the plant\ifmaize; it gives the best forecasts on both rice and maize\fi.
  \item 
 We show that standard metrics such as PSNR and SSIM are useless in guiding the generation of ecologically meaningful plants, and introduce an evaluation of identity, geometry, color, and timing against the plant's growth series.
\end{itemize}

\section{Related Work}
\label{sec:related}

\subsection{World Models in Computer Vision}
\label{sec:related-wm}

\paragraph{World models and long-horizon prediction.}
World models learn how an environment evolves, first as latent simulators for control~\cite{ha2018recurrent,hafner2023dreamerv3} and now as large video generators presented as world simulators~\cite{brooks2024sora,agarwal2025cosmos,wan2025wan} for games~\cite{bruce2024genie,valevski2024gamengen}, driving~\cite{hu2023gaia}, and robots~\cite{yang2024unisim,assran2025vjepa2}.
Longer horizons have been reached by temporal abstraction, which predicts directly at coarse or adaptive steps~\cite{neitz2018adaptive,gregor2019tdvae,jayaraman2019time,saxena2021clockwork,farebrother2026jumpy}, and learned weather models forecast physical fields days ahead~\cite{lam2023graphcast,bi2023pangu}.
These models primarily predict physical dynamics, including object motion, interactions, and the evolution of physical fields.
\emph{We open a regime they do not cover}: the growth of an organism over days to months, driven by a state that the image shows only indirectly.

\paragraph{Probing and evaluating world knowledge.}
Video generators often violate physical laws despite realistic frames~\cite{bansal2024videophy,kang2024howfar,motamed2026physicsiq} and lack a consistent time scale~\cite{wu2026pulse}.
Time-lapse generation~\cite{zhou2016learning,xiong2018learning,yuan2025magictime,yuan2024chronomagic} targets slow processes from text, with no individual's real future as ground truth, keyframe interpolation~\cite{wang2025inbetweening} is given the last frame, and face aging~\cite{orel2020lifespan} involves no new organs.
These tasks often use pixel and perceptual metrics~\cite{wang2004ssim,zhang2018lpips,oquab2024dinov2} to measure visual similarity, but such metrics can favor blurry averages under uncertainty~\cite{blau2018perception,denton2018svg}. In addition, FID~\cite{heusel2017fid} compares image distributions and reflects both visual fidelity and diversity, but does not separate their contributions to the score.
We extend this probing from physical consistency to the timing of biological development. Genotype, environment, and management shape a plant's developmental trajectory, so elapsed time alone does not determine its future stage. A forecaster must infer where the plant is in its development and how far it will progress over the requested horizon. A realistic image can still be the wrong forecast: its size, color, or organs may be plausible but mistimed. We therefore score timing by retrieval in the same plant's photo series (Sec. 3.2).

\paragraph{Reasoning before generation.}
Multimodal language models have been used to plan image edits~\cite{fu2024mgie,huang2024smartedit} and to reason before synthesis~\cite{fang2025got}, extending chain-of-thought prompting~\cite{wei2022cot}, and CausalWM~\cite{xu2026causalwm} reasons about causal variables before predicting video.
Their reasoning concerns layout, semantics, or short-term physics.
\emph{\clock differs in what it reasons about and where its supervision comes from}: it advances a hidden developmental state by a requested horizon and states it in image measurements, in a schema designed with agronomic knowledge.
Its training scripts are written backwards from real future images around recorded stages and measured values, in the spirit of privileged information~\cite{vapnik2009lupi,chen2019lbc} and rationalization~\cite{zelikman2022star}.

\subsection{Modeling Growth in Agriculture}
\label{sec:related-eco}

\paragraph{Process-based crop and plant models.}
Agronomists have long forecast crops with process-based models: crop models such as ORYZA2000~\cite{bouman2001oryza}, DSSAT~\cite{jones2003dssat}, APSIM~\cite{holzworth2014apsim}, and WOFOST~\cite{dewit2019wofost} advance a developmental state with temperature and day length and simulate biomass and yield from weather, soil, and structural plant models simulate architecture~\cite{prusinkiewicz1990algorithmic,vos2010fspm}.
They need cultivar- and site-specific parameters and output numbers or geometry rather than the appearance of a particular plant.
\emph{We keep their central idea, a developmental state advanced through time, but infer it from the photo and render it as an image}.

\paragraph{Observing and forecasting growth from images.}
Image-based phenotyping describes the plant's current state: vision models segment organs, measure traits, and estimate plant age from photographs~\cite{scharr2016leaf,yang2014combining,bhatt2025gromo}.
Growth forecasting goes further by predicting its later appearance.
Plant-growth GANs predict future images from earlier photos, using fixed prediction intervals or explicit time, treatment, and biomass conditions, and assess their outputs through image quality and trait agreement~\cite{drees2021temporal,drees2024datadriven,wang2025maize}.
Agricrafter~\cite{agricrafter} generates full-growth-cycle crop videos with diffusion models, and 4D reconstruction~\cite{luo2026growflow} recovers growth that has already been imaged.
We explicitly forecast the future developmental state of the plant to guide image generation from a current photo at horizons of up to 20 weeks, and evaluate the developmental timing of the plant's photo series\ifmaize, on two crops\fi.

\ifcombinedanalysis
\begin{figure}[t]
  \centering
  \includegraphics[width=\textwidth]{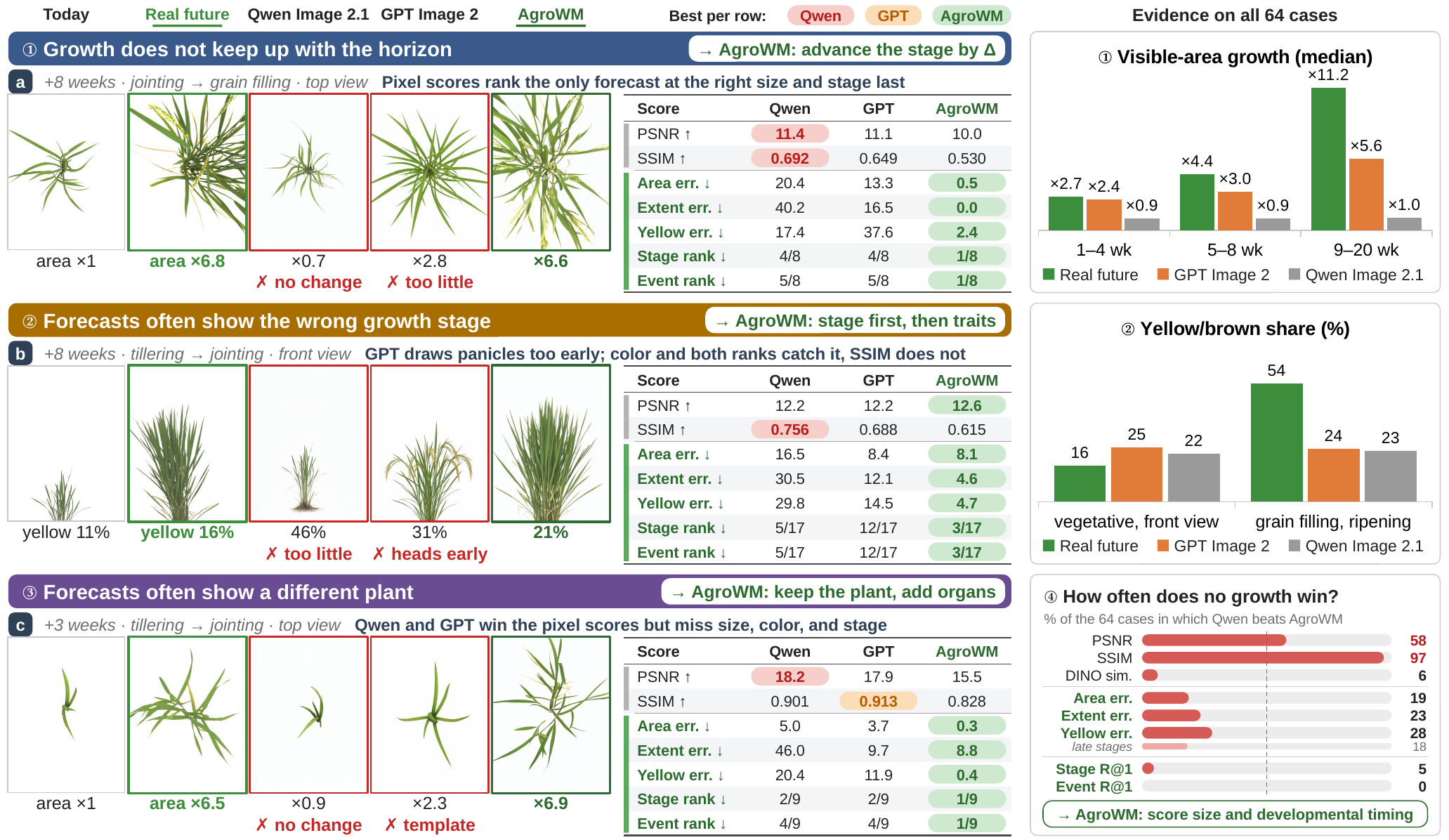}
  
  \caption{\textbf{How existing models fail to forecast growth, and why standard scores miss it.} 
  }
  \label{fig:analysis}\label{fig:metrics}
  
\end{figure}
\fi

\section{Preliminary Analysis}
\label{sec:analysis}

We first check how state-of-the-art image generators forecast growth and whether standard scores reflect their mistakes.
We test two of them: GPT Image 2, the general-purpose model with the highest Stage R@1 in \cref{tab:main}, and Qwen Image 2.1, the base of our \renderer.
The 64 validation cases come from 64 cultivars, half in each view,
with 22, 24, and 18 cases at 1--4, 5--8, and 9--20 rounded weeks
(the longest is 15 weeks; \cref{fig:analysis}).
Each receives today's photo and a prompt with the cultivar, view, and horizon, plus a keep-list that asks to keep the plant's identity and to avoid premature yellowing or panicles.

\subsection{How Existing Models Fail}
\label{sec:analysis-failures}

\paragraph{\ding{172} Growth does not keep up with the horizon.}
\label{sec:analysis-timeblind}\nobreak%
The further ahead a forecast looks, the more the plant should change, but neither generator keeps pace with the real change.
Qwen Image 2.1 barely changes the plant: its median projected area
is $0.93\times$ to $0.98\times$ today's, while real plants grow
$2.7\times$, $4.4\times$, and $11.2\times$ in the three ranges.
GPT Image 2 grows the plant with the horizon but falls further behind,
reaching $2.4\times$, $3.0\times$, and $5.6\times$
(\ifcombinedanalysis\cref{fig:analysis}a\fi).

\paragraph{\ding{173} Forecasts often show the wrong growth stage.}
How a plant looks depends on its growth stage, and the generators often draw the wrong one.
Although the prompt forbids it, GPT Image 2 yellows young plants early: in 8 of the 15 front-view cases that stay vegetative, the share of yellow or brown plant pixels in its forecast exceeds the real plant's by more than 10 percentage points (\ifcombinedanalysis\cref{fig:analysis}b\fi).
Both models keep ripening plants green: in the 22 cases at grain filling or ripening, the real plants are 54\% yellow or brown on average, against 24\% for GPT Image 2 and 23\% for Qwen Image 2.1 (\ifcombinedanalysis\ding{173} in \cref{fig:analysis}\fi).
The horizon alone cannot fix the stage: at 9--20 rounded weeks, 6 of the 18 plants have not yet headed, while 9 are filling grain or ripening, so each plant follows its own clock.

\paragraph{\ding{174} Forecasts often show a different plant.}
A forecast must show the same individual plant, but GPT Image 2 often changes its form toward a typical rice plant.
In front views it draws a taller, upright plant, more than 10 points of image height too tall in 13 of the 32 cases (\ifcombinedanalysis\cref{fig:supp-failures}e\fi); in top views it draws a symmetric rosette (\ifcombinedanalysis\cref{fig:analysis}a, c\fi).
Qwen Image 2.1 keeps the individual mostly by not changing it.

Together, these failures suggest that the generators jump from today's photo and the horizon straight to a new photo, without working out the plant's developmental state, and fall back on a generic picture of rice.
Our method is therefore built around three requirements: (T1)~advance the developmental state explicitly (\ding{172}); (T2)~read the plant's own stage first and derive its traits from it (\ding{173}); and (T3)~keep the individual and change only what development changes (\ding{174}).

\begin{figure}[t]
  \centering
  \includegraphics[width=\textwidth]{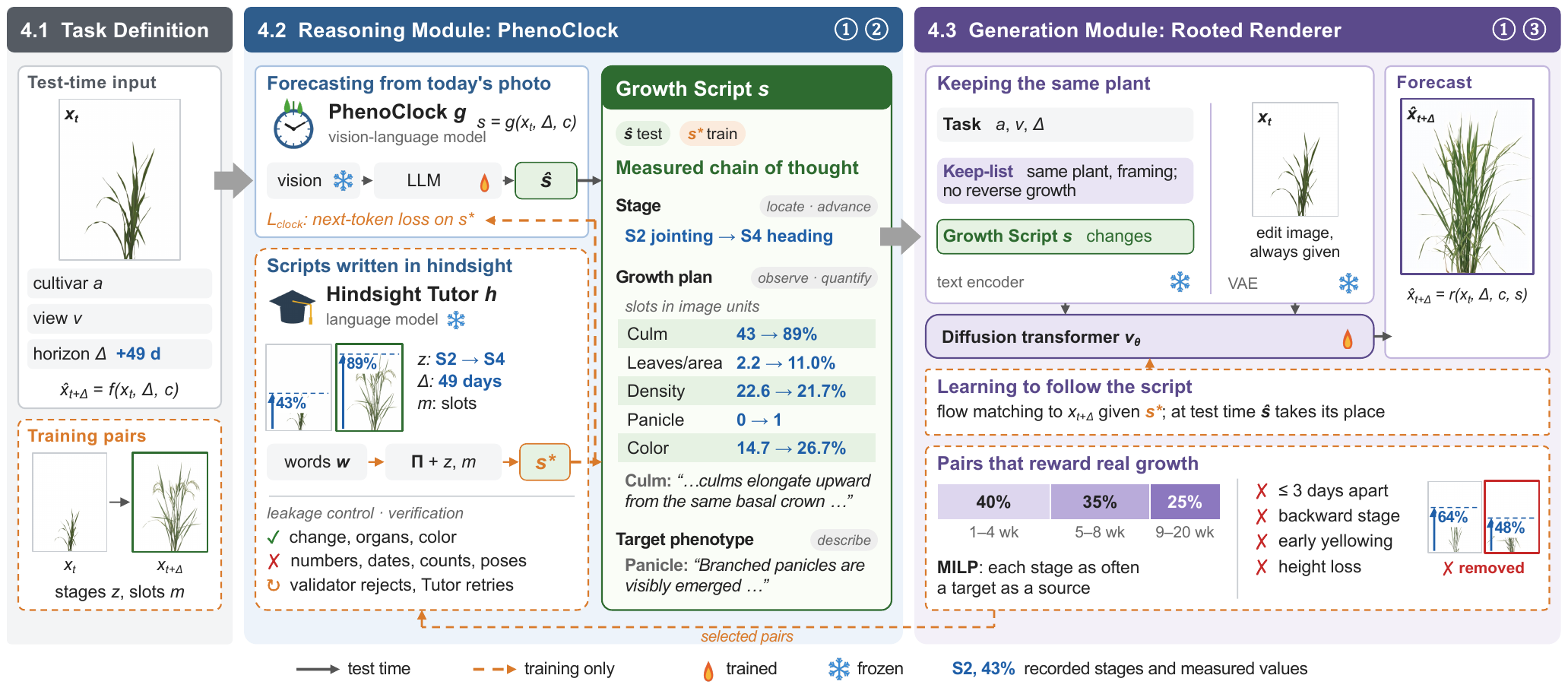}
  
  \caption{\textbf{\ours forecasts development before rendering growth.}
  \clock writes a \script with the future stage and targets for height, plant area, and color, and the \renderer edits the source photo to follow it.
  Example: a validation pair 49 days apart (7 weeks in the prompt), with the \tutor's reference script (center) and the forecast from \clock's own script (right).
  }
  \label{fig:pipeline}
  
\end{figure}

\subsection{Our Evaluation Design}
\label{sec:protocol}

Assessing whether forecasts meet these requirements calls for more than visual similarity. We first examine where standard scores fall short, then define an evaluation of identity, size, color, and developmental timing. 
\paragraph{\ding{175} Standard scores reward no growth.}
\label{sec:analysis-metrics}\nobreak%
Standard scores do not catch these failures: pixel scores favor an unchanged photo, and the other scores each see only part of the change.
In the 64 cases, plants cover only 14\% of the real future photo, so PSNR and SSIM mostly measure the white background, and an unchanged copy of today's photo gets the highest mean PSNR and SSIM of all forecasts we score (14.9\,dB and 0.81, against 14.2\,dB and 0.75 for GPT Image 2).
DINOv2 similarity~\cite{oquab2024dinov2} depends less on the background,
but its scale shifts with the horizon: the copy scores 0.74 at 1--4
rounded weeks and 0.41 at 9--20, so a value means little without its horizon.
Single traits see growth but not development: GPT Image 2 beats the copy and Qwen Image 2.1 in projected area in 52 of the 64 cases, yet in 33 of them the forecast looks most like a photo from another stage\ifcombinedanalysis\ (\cref{fig:analysis}b)\fi.

\paragraph{What must a forecast get right?}
We score four parts of a forecast separately.
\emph{(i)~\textbf{Identity}} (\ding{174}): DINOv2 feature similarity (DINO) and foreground-mask IoU with the real future.
\emph{(ii)~\textbf{Size and shape}} (\ding{172}, \ding{174}): errors, in points of the image, of projected area, plant height, and density, the plant's share of its convex hull.
\emph{(iii)~\textbf{Color}} (\ding{173}): error of the yellow--brown share of plant pixels.
\emph{(iv)~\textbf{Timing}} (\ding{172}, \ding{173}): we rank the photos of the same plant and view, today's included (14 on average), by DINOv2 ViT-B/14 similarity to the forecast; Stage R@1 and Event R@1 are the shares of forecasts whose top photo lies in the target stage or is the real target.
Since these photos share the plant and camera, a forecast that does not grow matches today's photo: Qwen Image 2.1 reaches an Event R@1 of only 1.1\%, worse than picking a photo at random (at least 7\%; \cref{tab:main}).
On the 2{,}000 validation queries, PSNR and SSIM prefer this no-growth forecast to \ours on 51\% and 93\%, the timing scores on 6\% and 1\% (\cref{tab:metric-validity}; \ifcombinedanalysis\ding{175} in \cref{fig:analysis}\fi), so PSNR and SSIM are for reference only.
A \script states the target size and color, so these errors may favor \ours; the timing scores use only the forecast image and recorded labels.

\section{Method}
\label{sec:method}

\ours answers two questions in turn, what the plant will become and how it will then look (\cref{fig:pipeline}), instead of asking one generator to guess the developmental state, the amount of growth, and what to keep all at once (\cref{sec:analysis}).
The reasoning module, \clock (\cref{sec:planner}), answers the first: from today's photo and the query, it predicts a measured \script (T1, T2).
The generation module, the \renderer (\cref{sec:renderer}), answers the second: it edits the same photo as the script says (T3).
In training, a \tutor that sees both photos of a pair writes the words of its reference script around recorded stages and measured values.

\subsection{Task Definition}
\label{sec:task-def}

The model sees one photo and predicts the plant at a requested future horizon.
Given a photo $x_t$ of a plant at time $t$, a horizon $\Delta$, and
conditions $c=(a,v)$, the cultivar $a$ and camera view
$v\in\{\text{front},\text{top}\}$, an agricultural world model
predicts the same plant at time $t+\Delta$:
\begin{equation}
  \hat{x}_{t+\Delta} = f(x_t, \Delta, c), \qquad 1 \le \Delta \le T,
  \label{eq:task}
\end{equation}
where $T$ is the rest of the season.
In the rice experiments, elapsed intervals are rounded to whole weeks
for the model input, while target photos retain their recorded observation dates.
The model must infer the current stage from the photo, since no stage label is given.
The recorded stages and slot values (\cref{tab:slots}) of a pair $(x_t, x_{t+\Delta})$ are available only in training and for evaluation.
\ours splits $f$ into its two modules,
\begin{equation}
  s = g(x_t, \Delta, c), \qquad \hat{x}_{t+\Delta} = r(x_t, \Delta, c, s),
  \label{eq:factorization}
\end{equation}
where \clock $g$ writes the \script $s$ and the \renderer $r$ edits the source photo accordingly.

\subsection{Reasoning Module: \clock}
\label{sec:planner}

\clock answers the first question, what the plant will become.
It is a vision-language model whose output, a structured chain of thought, spells out how the plant will develop.
Unlike free-form reasoning before generation~\cite{fang2025got}, this chain of thought must be exact enough to draw and to check, and WGSR does not provide it.
We therefore define what it contains, write training examples in hindsight, and teach \clock to produce them from today's photo.

\paragraph{Measured chain of thought.}
Like an agronomist's note, a \script states the stage first and the traits second, in three blocks of fixed order,
\begin{equation}
  s = \big( z_t,\, z_{t+\Delta};\; \{ (d_k,\, m^k_t,\, m^k_{t+\Delta}) \}_{k};\; \{ e_k \}_{k} \big),
  \label{eq:script}
\end{equation}
where $k$ runs over the slots $\mathcal{K}_v$ of view $v$: culm height, plant area, density or canopy, panicles, and color (\cref{tab:slots}).
The stage block gives today's and the future stage on eight stages S1--S8 adapted from the IRRI scale~\cite{irri2013ses} (tillering, jointing, booting, heading, grain filling, milk and dough, yellow ripening, and maturity); the growth-plan block gives, for each slot, a sentence $d_k$ on the visible change with today's and the target value; and the target-phenotype block describes the end state ($e_k$).
Read in order, a \script \emph{locates} the plant, \emph{advances} it by $\Delta$, and then \emph{observes}, \emph{quantifies}, and \emph{describes} each slot.
All values except the panicle label, which follows the cultivar's heading date, are image measurements, such as the plant's extent from the bottom edge and its share of the image: each is a target for \clock, a condition for the \renderer, and a quantity measured again on the forecast (\cref{sec:protocol}).

\paragraph{Scripts written in hindsight.}
We write the training scripts backwards, from pairs whose future we know.
A language model explains well but measures poorly, so the \tutor $h$ sees both photos and the facts of a pair, the recorded stages $z=(z_t,z_{t+\Delta})$, the interval, and the slot values $m=(m_t,m_{t+\Delta})$, but writes only the words $w$, in a strict schema; a fixed template $\Pi$ then inserts the stages and values and computes each change:
\begin{equation}
  w = h\big(x_t,\, x_{t+\Delta},\, \Delta,\, z,\, m\big), \qquad s^\star = \Pi(w,\, z,\, m).
  \label{eq:tutor}
\end{equation}
The stages and numbers of $s^\star$ are thus copied from the records and measurements.
Because $h$ sees the answer, its prompt allows words only on what a forecaster could know: the kind and direction of change, organ emergence consistent with the panicle value, and color progression, never numbers, dates, organ counts, or future poses (leakage control).
A validator checks the structure, the stages against the records, and forbidden forms such as numbers in the words, and $h$ rewrites rejected scripts (verification).

\paragraph{Forecasting from today's photo.}
At test time the future is unknown, so \clock must write the same \script from today's photo, the horizon, and the conditions alone.
We fine-tune \clock with a next-token loss on $s^\star$ whose tokens fall into three groups, the stage names, the metric fields, and the remaining script text:
\begin{equation}
  \mathcal{L}_{\text{clock}} = 0.1\,\mathcal{L}_{\text{stage}} + 0.4\,\mathcal{L}_{\text{metric}} + 0.5\,\mathcal{L}_{\text{text}},
  \label{eq:clock-loss}
\end{equation}
where each term is the loss $-\log p_g\big(s^\star_i \mid s^\star_{<i},\, x_t,\, \Delta,\, c\big)$ over the tokens $s^\star_i$ of one group.
The future reaches \clock only through this label, as in learning with privileged information~\cite{vapnik2009lupi,chen2019lbc}, and the label's source-side stage and values also teach \clock to read today's plant.

\subsection{Generation Module: \renderer}
\label{sec:renderer}

The \renderer answers the second question, how the plant will then look: it should add the organs and colors that the \script states and keep the rest of today's photo, whereas the generators of \cref{sec:analysis} either leave the plant unchanged (\ding{172}) or alter it (\ding{174}).
We build it from a pretrained image generation and editing model with three choices, one per paragraph below: it edits today's photo, it learns from pairs that reward real growth, and it trains on the \tutor's scripts.

\paragraph{Keeping the same plant.}
The \renderer edits today's photo instead of drawing a new one: the source photo is the edit image in training and at test time, and every target keeps the frame of its source, in which the numbers of the \script are defined.
The prompt states the task (accession, view, and horizon in whole weeks), a fixed keep-list (same identity, background, framing, and scale; development only forward, without premature yellowing or panicles), and the \script.
A prompt alone is not enough, since GPT Image 2 received the same keep-list and still altered the plant (\cref{sec:analysis}); the \renderer instead learns to follow it from its training pairs.

\paragraph{Pairs that reward real growth.}
A generator learns whatever its training pairs reward.
Among the candidate pairs, however, long horizons are scarce, short ones are nearly solved by copying, and early stages appear mostly as sources and late ones as targets.
We therefore fix a target share for each horizon range and, within each split, select candidate pairs within weekly quotas by a mixed-integer program that makes each stage appear about as often as a source as a target, with at least one pair per plant (\cref{eq:milp} in \cref{sec:supp-pairs}).
From the candidates, we drop pairs whose change is not growth, such as backward stages or early yellowing, and draw the final pairs stratified by source and target stage; the fine-tuned baselines use them too.

\paragraph{Learning to follow the script.}
The \renderer should learn that following the script pays off, so we train it with the \tutor's scripts $s^\star$, whose stages and values match the target, and give it the scripts $\hat{s}$ of \clock only at test time; imperfect predicted scripts in training could teach it to ignore the script.
We fit each triplet $(x_t, s^\star, x_{t+\Delta})$ with the flow-matching objective of the base model~\cite{lipman2023flow,liu2023rectified}:
\begin{equation}
  \mathcal{L}_{\text{ren}} = \mathbb{E}_{\tau,\epsilon} \big\| v_\theta\big(\xi_\tau, \tau;\, \mathcal{E}(x_t),\, p^\star\big) - (\epsilon - \xi_0) \big\|_2^2,
  \label{eq:ren-loss}
\end{equation}
where $\mathcal{E}$ is its VAE encoder, $\xi_0=\mathcal{E}(x_{t+\Delta})$, $\xi_\tau=(1-\tau)\,\xi_0+\tau\,\epsilon$ with $\epsilon\sim\mathcal{N}(0,I)$, and $p^\star$ is the prompt built from $(\Delta, c, s^\star)$; the text and VAE encoders stay frozen.

\section{Experiments}
\label{sec:exp}

We compare \ours with state-of-the-art image generators and domain-specific models\ifmaize{} on rice and maize\fi, test it at long horizons, and assess its modules and the accuracy of its \scripts.

\begin{table}[t]
  \centering
  \caption{\textbf{Main comparison on rice.} 
  Bold: best per column; PSNR and SSIM for reference only.
  }
  \label{tab:main}
  \fontsize{8}{10}\selectfont
  \setlength{\tabcolsep}{5pt}
  \begin{adjustbox}{max width=\textwidth}
  \begin{tabular}{@{}lcccccccccc@{}}
    \toprule
    & \multicolumn{2}{c}{Identity} & \multicolumn{3}{c}{Geometry} & Color & \multicolumn{2}{c}{Timing} & \multicolumn{2}{c}{Pixel (ref.)} \\
    \cmidrule(lr){2-3}\cmidrule(lr){4-6}\cmidrule(lr){7-7}\cmidrule(lr){8-9}\cmidrule(l){10-11}
    Method & DINO$\uparrow$ & IoU$\uparrow$ & Area$\downarrow$ & Height$\downarrow$ & Density$\downarrow$ & Yellow$\downarrow$ & \makecell{Stage\\R@1$\uparrow$} & \makecell{Event\\R@1$\uparrow$} & PSNR$\uparrow$ & SSIM$\uparrow$ \\
    \midrule
    \multicolumn{11}{l}{\emph{General-purpose}} \\
    GPT Image 2 & 0.709 & 0.303 & 5.59 & 11.49 & 6.31 & 24.40 & 35.3 & 12.8 & 13.23 & 0.689 \\
    GPT Image 2.5 & 0.721 & 0.299 & 6.92 & 12.57 & 6.70 & 29.17 & 34.8 & 10.8 & 12.92 & 0.694 \\
    Qwen Image 2.1 & 0.693 & 0.241 & 7.85 & 18.28 & 6.83 & 32.31 & 24.0 & 1.1 & \textbf{14.20} & 0.748 \\
    Qwen-Image-Edit-2511 & 0.568 & 0.188 & 28.84 & 20.46 & 25.08 & 33.76 & 26.5 & 9.2 & 9.54 & 0.476 \\
    FLUX.2 Klein & 0.496 & 0.259 & 41.58 & 22.47 & 36.51 & 34.11 & 25.2 & 9.6 & 7.36 & 0.392 \\
    FLUX 3 Image Klein Base & 0.683 & 0.244 & 13.87 & 17.65 & 12.38 & 35.14 & 24.1 & 1.4 & 12.98 & 0.691 \\
    LANCE & 0.636 & 0.240 & 23.81 & 20.90 & 24.17 & 31.61 & 24.5 & 3.8 & 10.55 & 0.591 \\
    \midrule
    \multicolumn{11}{l}{\emph{Domain-specific}} \\
    Copy & 0.702 & 0.252 & 7.17 & 19.06 & 6.24 & 29.33 & 23.2 & 0.0 & 14.02 & \textbf{0.750} \\
    LANCE FT & 0.815 & \textbf{0.342} & 3.91 & 8.04 & 5.62 & 15.95 & 46.9 & 16.9 & 13.94 & 0.709 \\
    FLUX.2 Klein FT & 0.767 & 0.305 & 4.70 & 11.00 & 6.03 & 24.59 & 39.3 & 18.0 & 13.91 & 0.697 \\
    Qwen Image 2.1 FT & 0.834 & 0.323 & 3.73 & 7.83 & 5.33 & 11.43 & 47.4 & 22.5 & 14.10 & 0.685 \\
    Agricrafter & 0.666 & 0.329 & 5.20 & 9.61 & 7.97 & 15.79 & 32.1 & 13.3 & 14.11 & 0.706 \\
    \midrule
    \multicolumn{11}{l}{\emph{Ours}} \\
    \ours{} & \textbf{0.860} & 0.327 & \textbf{3.51} & \textbf{7.43} & \textbf{4.98} & \textbf{10.00} & \textbf{54.6} & \textbf{24.8} & 14.14 & 0.686 \\
    \bottomrule
  \end{tabular}
  \end{adjustbox}
    
\end{table}

\subsection{Experimental Setup}
\label{sec:exp-setup}

\paragraph{Datasets.}
\label{sec:benchmark}
\ifmaize
We validate \ours on two crops that grow very differently, with two public datasets of the Open Plant Image Archive (OPIA)~\cite{cao2024opia}.
WGSR~\cite{tang2023wgsr} photographs 93 rice cultivars about once a week over the season; we keep its top view and one side (front) view with white backgrounds, labeled with the eight growth stages of \cref{sec:planner} and each cultivar's heading date.
WGSM~\cite{wang2023wgsm} photographs 224 maize inbred lines across nine growth stages, from vegetative growth to physiological maturity, and we keep one side view.
Each query pairs an earlier and a later photo of the same plant and view; rice pairs are 4 to 140 days apart (median 35).
We split by plant: rice has 20{,}000 training queries from 252 plants and 2{,}000 validation queries from 79 held-out plants, 78 of whose cultivars also occur in training, and maize has 6{,}400 and 1{,}600 queries from 348 and 84 plants (\cref{sec:supp-pairs}).
\fi

\paragraph{Baselines.}
The baselines fall into the two groups of \cref{tab:main}.
\emph{General-purpose} generators are state-of-the-art image generation and editing models: the closed-source GPT Image 2 and GPT Image 2.5 and the open-source Qwen Image 2.1, Qwen-Image-Edit-2511~\cite{wu2025qwenimage}, FLUX.2 Klein, FLUX 3 Image Klein Base~\cite{bfl2026flux3klein}, and LANCE~\cite{fu2026lance}.
Each receives today's photo and a prompt with the cultivar, view, and horizon, plus the \renderer's keep-list, without a \script (\cref{sec:supp-impl}).
\emph{Domain-specific} methods are built for this task.
Copy returns today's photo.
LANCE FT, FLUX.2 Klein FT, Qwen Image 2.1 FT, and Agricrafter~\cite{agricrafter},
a crop growth video generator, are fine-tuned on our data;
method-specific training details are given in \cref{sec:supp-impl}.

\paragraph{Implementation details.}
We fine-tune \clock from Qwen3.8-27B~\cite{qwen2026qwen38} with LoRA adapters~\cite{hu2022lora} and the \renderer from Qwen Image 2.1~\cite{qwen2026qwenimage21,wu2025qwenimage}, a generation and editing model with a 7B diffusion transformer, in all transformer weights; the \tutor is GPT-6.1 Sol~\cite{openai2026gpt61sol}.
Both modules are trained for two epochs on the 20{,}000 training queries with a global batch of 16.
At test time, the \renderer samples 20 denoising steps; the settings of all methods are in \cref{sec:supp-impl}.

\subsection{Main Results}
\label{sec:exp-main}

\ours outperforms all seven state-of-the-art generators on every identity, geometry, color, and timing score (\cref{tab:main}), with Stage and Event R@1 of 54.6\% and 24.8\%, against 34.8\% and 10.8\% for GPT Image 2.5 and 24.0\% and 1.1\% for Qwen Image 2.1.
These generators change the plant too little or without control: Qwen Image 2.1 barely changes it, the other open models miss its area by 14 to 42 points, two to six times the error of the copy, and GPT Image 2, the best of them in Stage R@1, reaches 35.3\%.
Fine-tuning on our pairs helps all three generators we fine-tune, raising Stage R@1 from 24.5\% to 46.9\% for LANCE, from 25.2\% to 39.3\% for FLUX.2 Klein, and from 24.0\% to 47.4\% for Qwen Image 2.1, all above the 32.1\% of Agricrafter.
Adding \scripts on top of the same backbone and pairs lifts \ours above every domain-specific model, by at least 7.2 and 2.3 points in Stage and Event R@1.
It also has the best identity and trait scores of all generated forecasts, except IoU, where LANCE FT and Agricrafter are slightly higher (0.342 and 0.329 against 0.327).

\ifmaize
\Needspace{17\baselineskip}
\subsection{Generalization to Maize}
\label{sec:exp-maize}

\begin{wraptable}{r}{0.47\textwidth}
  \centering
  \caption{\textbf{Generalization to maize.} 
  }
  \label{tab:maize}
  \small
  \setlength{\tabcolsep}{2pt}
  \begin{adjustbox}{max width=\linewidth}
  \begin{tabular}{@{}lcccc@{}}
    \toprule
    Method & DINO$\uparrow$ & Area$\downarrow$ & Height$\downarrow$ & Stage R@1$\uparrow$ \\
    \midrule
    Copy & 0.522 & 3.62 & 33.02 & 0.0 \\
    GPT Image 2 & 0.727 & 2.63 & 16.64 & 21.1 \\
    Qwen Image 2.1 FT & 0.764 & 1.97 & 11.12 & 30.1 \\
    \ours & \textbf{0.795} & \textbf{1.73} & \textbf{10.61} & \textbf{30.9} \\
    \bottomrule
  \end{tabular}
  \end{adjustbox}
  
\end{wraptable}

Maize tests the framework on a crop with a single stalk, tassels, and ears.
We train maize-specific models from the original pretrained backbones,
adapting the stage scale and organ descriptions
(\cref{sec:supp-pairs,sec:supp-impl}).
\ours has the best overall means on the four reported metrics
(\cref{tab:maize}): Stage R@1 is 30.9\%, against 30.1\% for
Qwen Image 2.1 FT and 21.1\% for GPT Image 2, with area and height
errors of 1.73 and 10.61 points.
Its Stage R@1 advantage over Qwen Image 2.1 FT is smaller than on rice.
\fi

\Needspace{10\baselineskip}
\subsection{Long-Horizon Prediction}
\label{sec:exp-time}

\begin{wrapfigure}{r}{0.46\textwidth}
  \centering
  \includegraphics[width=\linewidth]{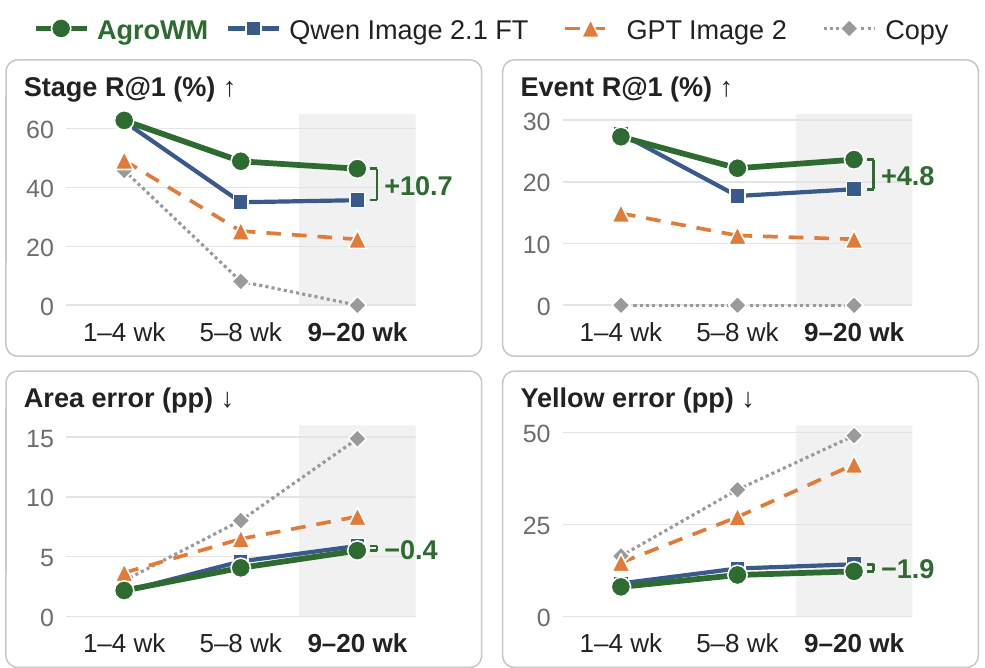}
  \caption{\textbf{Forecast quality by horizon on rice}. 
  }
  \label{fig:horizon}
\end{wrapfigure}

The further ahead a forecast looks, the less of the answer today's photo shows, so \cref{fig:horizon} splits the scores by horizon.
One to four weeks ahead, the stage often stays the same, so the copy of today's photo already reaches 45.8\% Stage R@1, and \ours and Qwen Image 2.1 FT are about equally accurate (Stage R@1 62.8\% and 62.4\%, Event R@1 27.3\% and 27.8\%).

\Needspace{20\baselineskip}

As the horizon grows, all four methods of \cref{fig:horizon} lose Stage R@1, \ours the least: from 1--4 to 9--20 weeks, it falls from 62.8\% to 46.4\%, against 62.4\% to 35.7\% for Qwen Image 2.1 FT and 49.2\% to 22.4\% for GPT Image 2.
From five weeks on, \ours is best on all four scores of \cref{fig:horizon}.
Its Stage R@1 lead over Qwen Image 2.1 FT is 0.4 points at 1--4 weeks but 13.9 and 10.7 points at 5--8 and 9--20 weeks, and at 9--20 weeks it also leads by 4.8 points in Event R@1 and lowers the area and yellow errors by 0.4 and 1.9 points.
Since the two share backbone, pairs, and schedule and differ in whether they use \scripts, reasoning about development helps most where today's photo leaves the future open.

\WFclear
\vspace{\baselineskip}
\Needspace{17\baselineskip}
\subsection{Ablation Study}
\label{sec:exp-ablation}

\begin{wraptable}{r}{0.47\textwidth}
  \centering
  \caption{\textbf{Ablation studies on rice.} R@1 in \%, area error in points.}
  \label{tab:ablation}
  
  \small
  \setlength{\tabcolsep}{2.5pt}
  \begin{adjustbox}{max width=\linewidth}
  \begin{tabular}{@{}lcccc@{}}
    \toprule
    Variant & \makecell{Stage\\R@1$\uparrow$} & \makecell{Event\\R@1$\uparrow$} & Area$\downarrow$ & DINO$\uparrow$ \\
    \midrule
    \ours (full) & 54.6 & 24.8 & 3.51 & 0.860 \\
    \midrule
    w/o \clock{} & 47.4 & 22.5 & 3.73 & 0.834 \\
    w/o \renderer & 24.8 & 12.3 & 8.22 & 0.565 \\
    \midrule
    \tutor (Upper Bound) & 58.9 & 28.6 & 2.92 & 0.874 \\
    \bottomrule
  \end{tabular}
  \end{adjustbox}
  
\end{wraptable}

We compare \ours with a script-free fine-tuning control, a script-only base generator, and a target-informed script reference in \cref{tab:ablation}.
Without \clock (Qwen Image 2.1 FT, trained and run without \scripts), Stage and Event R@1 drop by 7.2 and 2.3 points (95\% CI 5.2--9.2 and 0.5--4.2, paired bootstrap over plants), and Stage R@1 by 10.7 points at 9--20 weeks (\cref{sec:exp-time}).
Without the \renderer, the base Qwen Image 2.1 receives the \scripts of \clock in its prompt but neither the source photo nor fine-tuning: Stage and Event R@1 fall to 24.8\% and 12.3\%, the area error rises to 8.22 points, and DINO drops to 0.565, so a good script is not enough without a renderer that keeps the plant.
With the \tutor's target-informed scripts, the same \renderer reaches 58.9\% and 28.6\% (\cref{sec:exp-script}).

\Needspace{18\baselineskip}
\subsection{How Accurate Is the Growth Script?}
\label{sec:exp-script}

\begin{wraptable}{r}{0.47\textwidth}
  \centering
  \caption{\textbf{\script accuracy on the rice validation split}, against the recorded stages and measured slots of the real future (height: 1{,}000 front views). $\pm1$: within one stage. 
  Similarity: mean cosine similarity of BGE~\cite{xiao2024cpack} embeddings to the \tutor's 18{,}000 slot descriptions.
  }
  \label{tab:script}
  
  \small
  \setlength{\tabcolsep}{3pt}
  \begin{adjustbox}{max width=\linewidth}
  \begin{tabular}{@{}lccc@{}}
    \toprule
    & \textit{Current} & \textit{Future} & \textit{Future $\pm1$} \\
    Stage accuracy (\%) $\uparrow$ & 89.90 & 78.90 & 99.40 \\
    \midrule
    & \textit{Height} & \textit{Area} & \textit{Yellow} \\
    Target MAE (pp) $\downarrow$ & 4.81 & 2.16 & 8.24 \\
    Change-direction Macro-F1 $\uparrow$ & 0.659 & 0.708 & 0.556 \\
    \midrule
    Semantic similarity (cosine) $\uparrow$ & \multicolumn{3}{c}{0.842} \\
    \bottomrule
  \end{tabular}
  \end{adjustbox}
  
\end{wraptable}

Errors in the \script carry over to the forecast, so we check the scripts of \clock against the recorded stages and measured traits of the real future (\cref{tab:script}).
\clock names today's stage correctly in 89.9\% of the queries and the future stage in 78.9\%, and 99.4\% of its future stages are at most one stage off, so advancing the stage is harder than reading it.
Its target values are off by 4.81, 2.16, and 8.24 points for height, area, and yellow, and 1{,}999 of its 2{,}000 scripts pass the strict format check.
These errors cost 4.3 points of Stage R@1: with the \renderer fixed, the \tutor's scripts reach 58.9\% and those of \clock 54.6\% (95\% CI of the gap 2.2--6.3; \cref{tab:ablation}), so better scripts would give better forecasts.

\section{Conclusion and Future Work}
\label{sec:conclusion}

We asked whether world models can forecast the slow change of living things, and found that two state-of-the-art image generators either barely change a rice plant or change it at the wrong pace and stage, while pixel scores favor the unchanged plant.
\ours{} narrows this gap by reasoning before rendering: it first forecasts a \script of the plant's development, learned from scripts written in hindsight, and then edits the photo to match.
Reasoning adds 7.2 points of Stage R@1 over the same backbone without \scripts, most beyond four weeks, and puts \ours ahead of state-of-the-art generators and domain-specific models\ifmaize{} on both rice and maize, two crops with different body plans and stage scales\fi.
Because each forecast is a photo of the same plant, we can check its identity, size, color, and developmental timing against the plant's own later photos.
The rice results come from potted plants on one greenhouse platform, with the background removed, mostly of cultivars seen in training, and from pairs filtered for growth.
Future work will condition forecasts on weather and management records and test them on new cultivars, more species, and field images.

\bibliographystyle{unsrt}
\bibliography{main}

\clearpage

\appendix
\section*{Supplementary Material}
\addcontentsline{toc}{section}{Supplementary Material}
\renewcommand{\thefigure}{S\arabic{figure}}\renewcommand{\theHfigure}{S\arabic{figure}}\setcounter{figure}{0}
\renewcommand{\thetable}{S\arabic{table}}\renewcommand{\theHtable}{S\arabic{table}}\setcounter{table}{0}
\makeatletter   %
\setlength{\@fptop}{0pt}\setlength{\@dblfptop}{0pt}
\makeatother
\newcommand{\metriccase}[1]{\ifcombinedanalysis\cref{fig:supp-metrics}#1\fi}

\ifcombinedanalysis
\begin{center}
  \captionsetup{type=figure}
  \includegraphics[width=\textwidth]{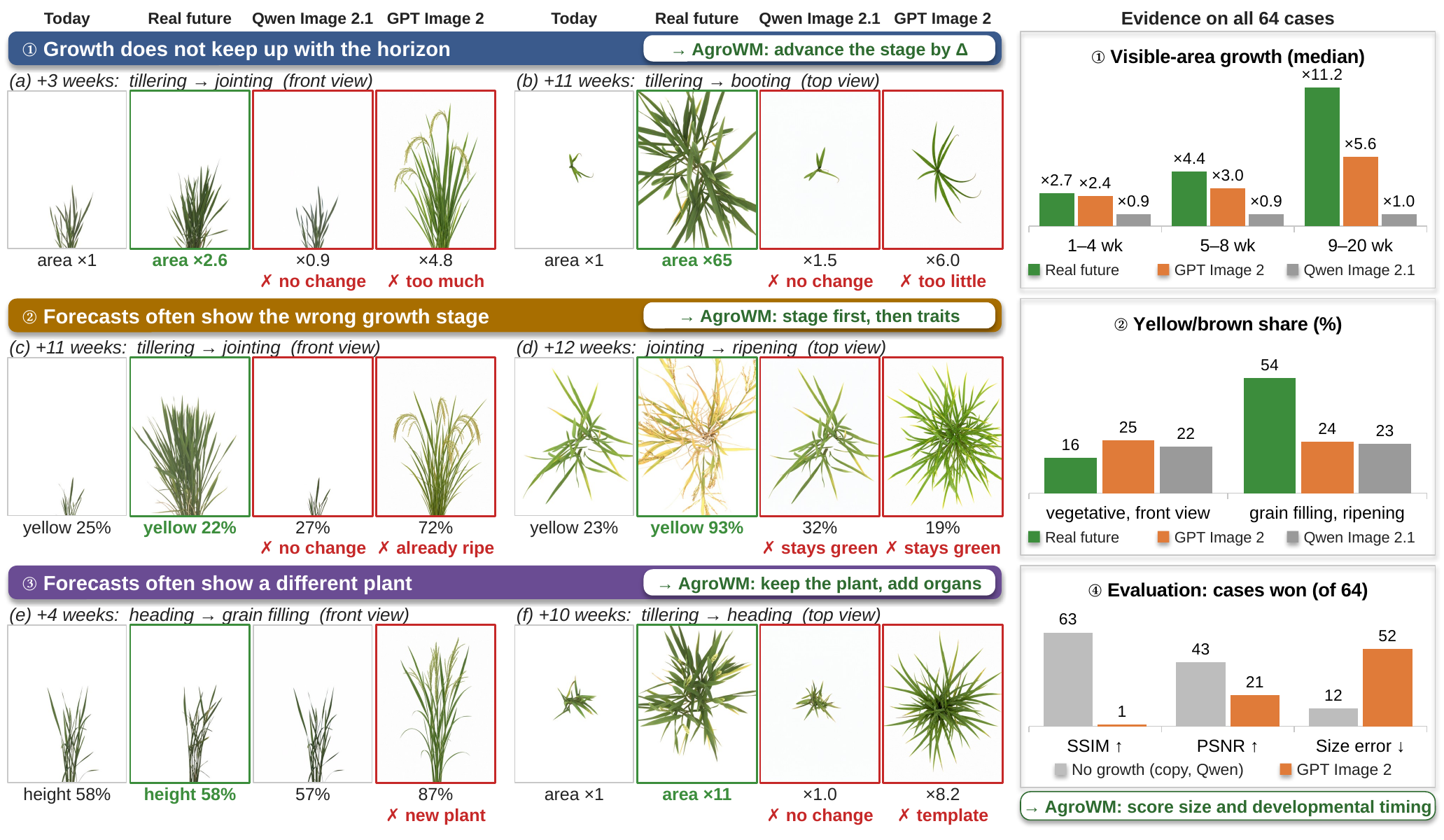}
  \captionof{figure}{\textbf{How existing models fail to forecast growth, two single cases per failure, front and top views.} Red frames mark failures, and the images of a case share one crop. Right: statistics over all 64 cases; \ding{175} counts the cases in which each score ranks no growth or GPT Image 2 best.}
  \label{fig:supp-failures}
\end{center}

\section{More Cases of the Preliminary Analysis}
\label{sec:supp-analysis}

\Cref{fig:analysis} in \cref{sec:analysis} shows one case per failure, chosen so that \ours is best on every trait error and retrieval rank shown.
In its top-view rows (a) and (c), PSNR and SSIM also prefer Qwen Image 2.1 or GPT Image 2, and this holds in only 3 of the 64 cases; in its front-view row (b), one of only 2 of the 32 front views in which every plant and timing score prefers \ours, SSIM prefers Qwen Image 2.1 while PSNR prefers \ours.
\Cref{fig:supp-failures} shows two cases per failure, chosen for the failure alone and covering both views, including the front-view failures that \cref{sec:analysis-failures} reports as statistics.
\Cref{fig:supp-metrics} shows all three, rows (a) and (c) of \cref{fig:analysis} and one front view, with the score labels in full and the share of cases in which each score prefers no growth, grouped as in \cref{tab:metric-validity}.

The cases of \cref{fig:supp-failures} add details that \cref{sec:analysis-failures} gives only as statistics.
GPT Image 2 paces its edits only loosely by the interval: after three weeks, it adds panicles to a plant that is still jointing (\cref{fig:supp-failures}a), whereas after eleven weeks, it keeps a young tillering plant at $6\times$ its area, against $65\times$ for the real plant (\cref{fig:supp-failures}b).
In front views, it turns a plant that is still jointing 72\% yellow or brown, against 22\% in reality (\cref{fig:supp-failures}c), and draws a plant taller, reaching 87\% of the image height where the real plant stays at 58\% (\cref{fig:supp-failures}e).
Ripening plants stay green in both models (\cref{fig:supp-failures}d), most of all in top views, where the real plants are on average 63\% yellow or brown, against 9\% for GPT Image 2.
Qwen Image 2.1 keeps the plant mostly by not changing it, but in 4 of the 64 cases it replaces the plant with a single culm or a field texture.

\begin{figure}[t]
  \centering
  \includegraphics[width=\textwidth]{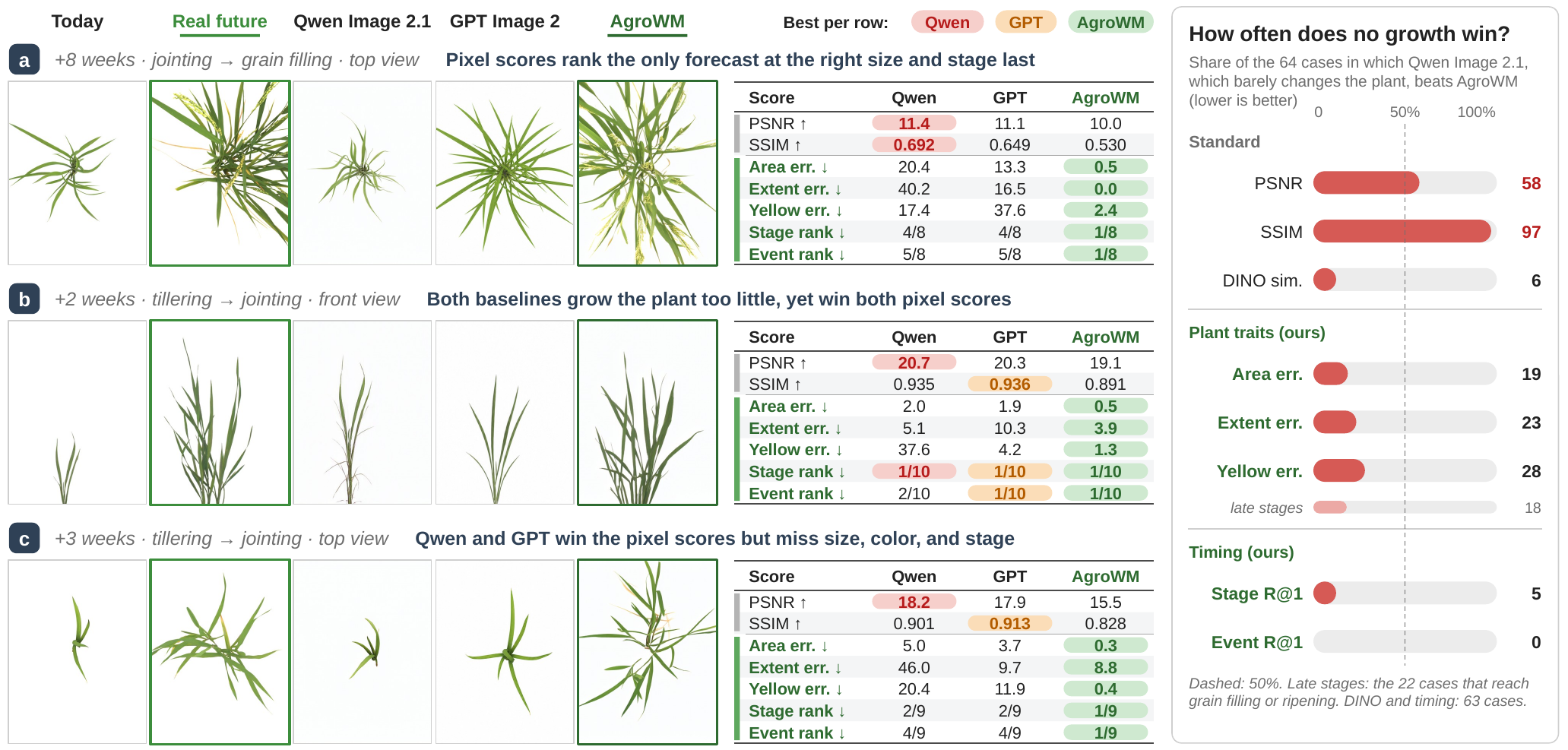}
  \caption{\textbf{What each score sees.} The three cases in which PSNR and SSIM prefer Qwen Image 2.1 or GPT Image 2 while \ours is best or tied on every plant and timing score; (a) and (c) are rows (a) and (c) of \cref{fig:analysis}.}
  \label{fig:supp-metrics}
\end{figure}
\else

\fi

\section{Analysis of Each Score}
\label{sec:supp-scores}

This section expands \cref{sec:protocol}.
We compare two forecasts with little or no growth, the exact copy of today's photo and Qwen Image 2.1, which barely changes it, with that of \ours, which sees only today's photo, the horizon, and the conditions: \cref{tab:metric-validity} counts how often each score prefers no growth over the 2{,}000 validation queries, and the 64 cases below show why.
All scores are computed by one evaluator on the white-background images.

\Needspace{28\baselineskip}
\begin{wraptable}{r}{0.49\textwidth}
  \centering
  \caption{\textbf{Which scores reward no growth?} Left: share of the validation queries on which a score strictly prefers the copy or Qwen Image 2.1 to \ours; ties count in the denominator but not as wins. Right: validation means (errors in percentage points, R@1 in \%). Height on 1{,}000 front views, DINO and R@1 on 1{,}964 queries, the rest on 2{,}000.}
  \label{tab:metric-validity}
  \small
  \setlength{\tabcolsep}{3pt}
  \renewcommand{\arraystretch}{1.1}
  \begin{adjustbox}{max width=\linewidth}
  \begin{tabular}{@{}llcccc@{}}
    \toprule
    & & \multicolumn{2}{c}{No growth wins (\%)} & \multicolumn{2}{c}{Validation mean} \\
    \cmidrule(lr){3-4}\cmidrule(l){5-6}
    & Score & Copy & Qwen & Qwen & \ours \\
    \midrule
    \multirow{3}{*}{\rotatebox{90}{\footnotesize Std.}}
    & PSNR $\uparrow$ & 45.5 & 50.8 & 14.20 & 14.14 \\
    & SSIM $\uparrow$ & 89.9 & 93.1 & 0.748 & 0.686 \\
    & DINO sim.\ $\uparrow$ & 25.7 & 18.7 & 0.693 & 0.860 \\
    \midrule
    \multirow{4}{*}{\rotatebox{90}{\footnotesize Traits}}
    & Area err.\ $\downarrow$ & 31.3 & 25.6 & 7.85 & 3.51 \\
    & Height err.\ $\downarrow$ & 29.6 & 29.1 & 18.28 & 7.43 \\
    & Density err.\ $\downarrow$ & 45.0 & 41.2 & 6.83 & 4.98 \\
    & Yellow err.\ $\downarrow$ & 28.8 & 21.3 & 32.31 & 10.00 \\
    \midrule
    \multirow{2}{*}{\rotatebox{90}{\footnotesize Time}}
    & Stage R@1 $\uparrow$ & 5.4 & 5.6 & 24.0 & 54.6 \\
    & Event R@1 $\uparrow$ & 0.0 & 0.6 & 1.1 & 24.8 \\
    \bottomrule
  \end{tabular}
  \end{adjustbox}
\end{wraptable}

\paragraph{Background agreement can dominate pixel scores.}
Over the 64 cases, the copy has a mean PSNR of 14.9\,dB and a mean SSIM of 0.81, against 14.7\,dB and 0.78 for Qwen Image 2.1 and 14.2\,dB and 0.75 for GPT Image 2.
Plants cover on average 14\% of a real future photo in the 64 cases, and 17\% over the validation set; the rest is white background.
PSNR and SSIM therefore weigh mostly pixels that a forecast leaves untouched when it does not grow, and they penalize a leaf drawn slightly off twice, where it is and where it should be, but a missing leaf only once.
SSIM prefers the copy to \ours in 63 of the 64 cases, and PSNR in 38.
In \metriccase{a}, the only forecast with the right size and stage receives the lowest PSNR and SSIM of the three, 1.3\,dB below Qwen Image 2.1.
Over the 2{,}000 validation queries, Qwen Image 2.1, which barely changes the plant, has the highest PSNR of all deployable models in \cref{tab:main}, only the copy has a higher SSIM, and both are above \ours.

\paragraph{Global features are not anchored in time.}
Cosine similarity of DINOv2 features~\cite{oquab2024dinov2} between the forecast and the real future depends little on the background and prefers the copy to \ours in only 8 of 63 cases.
Its value, however, depends on the horizon as much as on the forecast: the copy scores 0.74 at 1--4 weeks and 0.41 at 9--20 weeks, so the same value can mean a good long-horizon forecast or no change at a short horizon.
Over the validation set, the copy reaches 0.702, higher than five of the seven general-purpose generators and below only GPT Image 2 and GPT Image 2.5 (0.709 and 0.721).
A single similarity also cannot say in which direction a forecast is wrong, whether too early, too late, or not the same plant.

\paragraph{Plant traits measure growth, but no single trait measures time.}
Measured on the plant pixels alone, the errors in visible area, vertical extent, and yellow--brown share see the growth that pixel scores miss: in \metriccase{a} Qwen Image 2.1 is 20.4 and 40.2 points off in area and extent, against 0.5 and 0.0 for \ours.
Each trait, however, sees only one side of development.
A forecast of the right size can be at the wrong stage: GPT Image 2 has a smaller area error than the copy and Qwen Image 2.1 in 52 of the 64 cases (5.4 points on average over the 64 cases, against 10.0 for the copy), yet 33 of these 52 forecasts miss the target stage under Stage R@1.
Size cannot tell a green plant from a ripening one: in a top-view case twelve weeks ahead, GPT Image 2 misses the real area by only 3.6 points, but its plant is green where the real one is 94\% yellow and brown.
Color, in turn, is informative only when color changes: over all 64 cases Qwen Image 2.1 beats \ours on yellow--brown error in 18, because most real futures are still green, but in only 4 of the 22 cases that reach grain filling or ripening, such as \metriccase{a}, where it is 17.4 points off and \ours 2.4.
In front views, every plant reaches the bottom edge (all 64 source and target images), so the vertical extent equals the culm slot of the \script there.

\paragraph{Timing scores and no growth.}
The 36 of 2{,}000 queries whose plant has fewer than five photos in the view are skipped.
Since the gallery contains today's photo, the copy has an Event R@1 of zero, and its Stage R@1 is one only where the stage does not change, which is then the right answer.
Accordingly, Event R@1 never prefers the copy to \ours, and Stage R@1 does so in 3 of 63 cases, all with an unchanged stage.

\section{Slots of a Growth Script}
\label{sec:supp-slots}

\Cref{tab:slots} lists the slots of a \script (\cref{sec:planner}) and how each is measured.

\begin{table}[t]
  \centering
  \caption{\textbf{Slots of a \script}, image proxies measured on the plant pixels of the white-background photo (F: front, T: top view). Bottom: one growth-plan line of a \tutor script, validation pair, tillering to jointing in 27 days.}
  \label{tab:slots}
  \footnotesize
  \setlength{\tabcolsep}{2.5pt}
  \begin{tabularx}{\linewidth}{@{}lcXl@{}}
    \toprule
    Slot & View & Definition & Unit \\
    \midrule
    Culm & F & plant height (bottom edge to top plant pixel) & \% of image height \\
    Leaves/area & F, T & all plant pixels / image pixels & \% of image \\
    Plant density & F & plant pixels / pixels of their convex hull & \% \\
    Canopy & T & plant pixels / pixels of their convex hull & \% \\
    Panicle & F, T & on or after the accession's heading date & 0/1 \\
    Color development & F, T & yellow and brown / plant pixels, HSV thresholds & \% \\
    \midrule
    \multicolumn{4}{@{}p{\linewidth}@{}}{\textit{Culm: Culms extend upward from the same basal crown as the established tillers progress from short basal growth toward jointing-stage elongation. Metric = Height; Source = 23.44\%; Target = 62.39\%; Change = +38.95 pp.}} \\
    \bottomrule
  \end{tabularx}
\end{table}

\Cref{tab:script} in \cref{sec:exp-script} scores the \scripts of \clock against the recorded stages and measured slots of the real future.
Most of its future-stage errors are off by one stage, its direction of change is most often right for area and least for yellow, its descriptions read much like the \tutor's, and the one script that fails the strict format check states a height change that does not match its values and is passed to the \renderer as written.

\paragraph{Change-direction Macro-F1.}
We classify each predicted and reference source-to-target change as decrease ($d<-1$\,pp), stable ($-1\leq d\leq 1$\,pp), or increase ($d>1$\,pp), where $d$ is the change in the corresponding image proxy.
For each of height, area, and yellow--brown share, Macro-F1 is the unweighted mean of the F1 scores of these three classes.

\paragraph{Description similarity.}
We align the predicted and \tutor descriptions by view, section (Growth plan or Target phenotype), and slot, and remove the numeric evidence suffixes before comparison.
The 2{,}000 validation queries contain 18{,}000 reference descriptions across these two sections: five slots per front view and four per top view.
We encode the descriptions with \texttt{BAAI/bge-small-en-v1.5}~\cite{xiao2024cpack}, use L2-normalized CLS embeddings, and report the mean cosine similarity over matched descriptions.
This score measures agreement with the \tutor's wording and meaning; the stage and numeric scores assess factual accuracy.

\section{Datasets, Cleaning, and Training Pairs}
\label{sec:supp-pairs}

WGSR~\cite{tang2023wgsr}, a public dataset of the Open Plant Image Archive~\cite{cao2024opia}, holds 16{,}758 RGB photos of 93 rice accessions, taken from two sides and from the top about once a week from 42 to 182 days after sowing, on a LemnaTec phenotyping platform in a greenhouse in Beijing.
We keep the top view and one side view, called the front view, and generate white-background versions with Qwen Image 2.1~\cite{qwen2026qwenimage21}, whose prompt, rewritten per photo by its prompt enhancer, asks to keep the plant unchanged and to replace everything else with white, so that methods and scores see only the plant.
Some outputs keep remnants of the pot or rails or shift the color of young leaves; the color-coverage gate and the pair gates below remove part of them.
Each photo carries one of the eight growth stages of \cref{sec:planner}, and each accession its heading date.
To test on unseen plants, one random eligible plant per accession goes to validation before pairs are filtered and selected, and all photos of a plant stay in one split.
Stage-stratified sampling yields 10{,}000 training and 1{,}000 validation pairs, each in both views: 20{,}000 training queries from 252 plants of 85 accessions and 2{,}000 validation queries from 79 plants, one from each of 79 accessions, 78 of which also occur in training.
These queries use 7{,}540 photos of 331 plants from 86 accessions.

Before pairing, we drop the photos recorded as abnormal growth (three accessions and 57 single photos), all photos from the day a plant first received an artificial support, and a few others by hand.
Within each split, the candidate selection of \cref{sec:renderer} solves
\begin{equation}
  \min_{\mathbf{y}\in\{0,1\}^{P}} \textstyle\sum_{z} \big| N^{\text{src}}_{z} - N^{\text{tgt}}_{z} \big| + \lambda \textstyle\sum_{z} \big( | N^{\text{src}}_{z} - \bar{N}_{z} | + | N^{\text{tgt}}_{z} - \bar{N}_{z} | \big),
  \label{eq:milp}
\end{equation}
subject to weekly quotas and at least one pair per plant, where $\mathbf{y}$ selects among $P$ candidates, $N^{\text{src}}_{z}$ and $N^{\text{tgt}}_{z}$ count selected sources and targets in stage $z$, $\bar{N}_{z}$ is the count under the pool's stage proportions, and $\lambda$ weighs the second term.
It targets 40\%, 35\%, and 25\% of the pairs at 1--4, 5--8, and 9--20 weeks; after the gates and the stratified sampling, 46\%, 33\%, and 21\% of the validation queries fall in these ranges, so 54.4\% are beyond four weeks.
The gates drop pairs whose change is not growth: photos at most three days apart; backward stages or panicles; early yellowing, that is, a yellow--brown share of at least 0.35 in tillering or jointing, or a rise of at least 0.20 from these stages within two stages; a front-height loss of at least 5 points toward a target before yellow ripening; and color coverage below 70\%.
We attribute such changes to background removal, lighting, or leaning plants rather than growth; a pair that kept them would teach the \renderer to draw the artifact and the \tutor to explain it.

\ifmaize
WGSM~\cite{wang2023wgsm}, also available through OPIA, contains images of 224 maize inbred lines acquired on a high-throughput phenotyping platform across nine growth stages, which cover vegetative growth, tasseling, silking, grain filling, and physiological maturity.
We retain one side view and construct strictly forward pairs within each plant; the 768$\times$1{,}270 source images are padded by 10 pixels at the top to a 768$\times$1{,}280 canvas, then processed to a white background with connected-component cleanup.
The resulting queries cover 432 plants, split by plant into 348 training and 84 held-out validation plants, with 6{,}400 and 1{,}600 queries (\cref{sec:exp-maize}).
The protocol maps stages S1--S9 to V3, V5, V7, V10, V14, VT, R1, R3, and R6, respectively, and uses the unrounded interval in days; the Planner receives the plant ID, source image, and interval, but not the recorded query stages or future image.
\fi

\section{Implementation Details}
\label{sec:supp-impl}

For rice, photos are $672{\times}896$ pixels (width${\times}$height) in the front view and $896{\times}672$ in the top view.
The candidate selection of \cref{eq:milp} uses $\lambda=0.02$; the \tutor runs with reasoning effort xhigh, and all 22{,}000 queries end with a \tutor script that passes the validator.
\Cref{tab:supp-training} summarizes the training recipes for the learned methods reported in the paper.

\begin{table}[t]
  \centering
  \caption{\textbf{Training configurations for paper methods.} All local Planner and Executor runs use bfloat16. The image-model schedules use a peak learning rate followed by the listed warmup, stable, and cosine-decay fractions.}
  \label{tab:supp-training}
  \small
  \setlength{\tabcolsep}{4pt}
  \renewcommand{\arraystretch}{1.12}
  \begin{tabularx}{\linewidth}{@{}>{\raggedright\arraybackslash}p{2.0cm}>{\raggedright\arraybackslash}p{3.7cm}>{\raggedright\arraybackslash}p{2.8cm}>{\raggedright\arraybackslash}X@{}}
    \toprule
    Model / arm & Initialization and trainable components & Training data & Recipe \\
    \midrule
    Rice Planner & Qwen3.8-27B; LoRA $r=32$, $\alpha=64$; all linear layers; vision tower and aligner frozen & 20{,}000 queries; global batch 16 & 2 epochs, 2{,}500 updates; LR $10^{-4}$, 3\% warmup, cosine to $10^{-6}$; grouped loss $0.1/0.4/0.5$ \\
    Rice Simple Executor (Qwen Image 2.1 FT) & Original Qwen Image 2.1; full DiT; text and VAE encoders frozen; no \script & 20{,}000 queries; global batch 16 & 2 epochs, 2{,}500 updates; peak LR $10^{-5}$; 10\% warmup, 60\% stable, 30\% cosine to $10^{-6}$ \\
    Rice CoT Executor (\renderer) & Original Qwen Image 2.1; full DiT; text and VAE encoders frozen; teacher \scripts in training & 20{,}000 queries; global batch 16 & 2 epochs, 2{,}500 updates; peak LR $10^{-5}$; 10\% warmup, 60\% stable, 30\% cosine to $10^{-6}$; 8 GPUs \\
    Rice FLUX.2 Klein FT / LANCE FT & Corresponding pretrained image model; fine-tuned image-generation backbone; no Planner or \script & 20{,}000 queries; global batch 16 & 2 epochs, 2{,}500 updates; LR $10^{-5}$; 10\% warmup, 60\% stable, 30\% cosine to $10^{-6}$; 8 GPUs \\
    Agricrafter & DynamiCrafter\_512 initialization; UNet and image projector trained & 2{,}108 nine-observation windows from 178 rice training plants; global batch 4 & 6{,}000 updates; white letterbox canvas $320{\times}512$ \\
    Maize Planner & Qwen3.8-27B; LoRA $r=32$, $\alpha=64$; all linear layers; vision tower and aligner frozen & 6{,}400 queries; global batch 16 & 2 epochs, 800 updates; LR $10^{-4}$, 3\% warmup, cosine to $10^{-6}$; grouped loss $0.1/0.4/0.5$ \\
    Maize Simple / CoT Executors & Separately initialized Qwen Image 2.1 full DiTs; text and VAE encoders frozen; no script / teacher script in training & 6{,}400 queries per arm; global batch 16 & 2 epochs, 800 updates; peak LR $10^{-5}$; 10\% warmup, 60\% stable, 30\% cosine to $10^{-6}$ \\
    \bottomrule
  \end{tabularx}
\end{table}

At test time, the \renderer and Qwen Image 2.1 FT sample 20 denoising steps with a guidance scale of 1 in bf16, at the source size.
Among the open general-purpose baselines, Qwen Image 2.1, Qwen-Image-Edit-2511, and the 9B FLUX.2 Klein base model use 20 denoising steps in bf16, with true classifier-free guidance 1.0, true guidance 4.0, and guidance 4.0, respectively; Qwen Image 2.1 outputs are composited on white.
The 3B LANCE uses its official image-editing pipeline with 20 steps.
GPT Image 2 is called through its image-editing API with automatic quality, and its outputs are resized to the source size with Lanczos filtering.
GPT Image 2.5 and FLUX 3 Image Klein Base are included as evaluated in the main table; settings not present in the run records are not inferred here.
The rice image generators receive the keep-list of the \renderer's prompt and the horizon in rounded whole weeks.
All forecasts are scored at the source size.
For FLUX 3 Image Klein Base, only 998 queries have a valid height and 1{,}996 a valid density and yellow share.

\section{Baseline Experiments}
\label{sec:supp-baselines}

This section describes the baselines reported in the paper's main comparison\ifmaize{} and maize comparison (\cref{tab:maize})\fi.
The copy returns the source image unchanged.
The general-purpose image generators receive the source image and the task prompt described in \cref{sec:exp-setup}, including the accession, view, horizon, and the \renderer keep-list, without a predicted \script.
The open-model inference settings and output-size handling are given above; generated images are evaluated at the source dimensions.

The rice Qwen Image 2.1 FT, FLUX.2 Klein FT, and LANCE FT baselines are trained on the same rice training split without a Planner or \script; their training recipes are listed in \cref{tab:supp-training}.
Qwen Image 2.1 FT is the Simple Executor control for the \renderer.
The GPT Image API baselines and the copy do not use these training updates.

Agricrafter is a video-generation baseline adapted to WGSR, rather than an image-pair fine-tune; its window construction and training recipe are listed in \cref{tab:supp-training}.
At inference it predicts nine frames with DDIM for 50 steps and recursively advances by at most eight observation steps; the target frame is selected by the recorded observation-index difference, not by a requested number of days or weeks.

\ifmaize
The maize comparison contains the source copy, GPT Image 2, a Simple Qwen Image 2.1 Executor, and AgroWM.
The Planner, Simple Executor, and CoT Executor are each trained separately from their original pretrained model on the maize split; no rice-fine-tuned checkpoint is transferred.
Their configurations are listed in \cref{tab:supp-training}.
The CoT Executor is trained on teacher scripts but receives only the source-only Planner prediction at inference; GPT Image 2 uses the same source image and Simple task prompt, while the copy is unchanged.
\fi

\section{Planner Output Templates}
\label{sec:supp-cot-templates}

\begin{figure}[p]
  \centering
  \includegraphics[width=\textwidth]{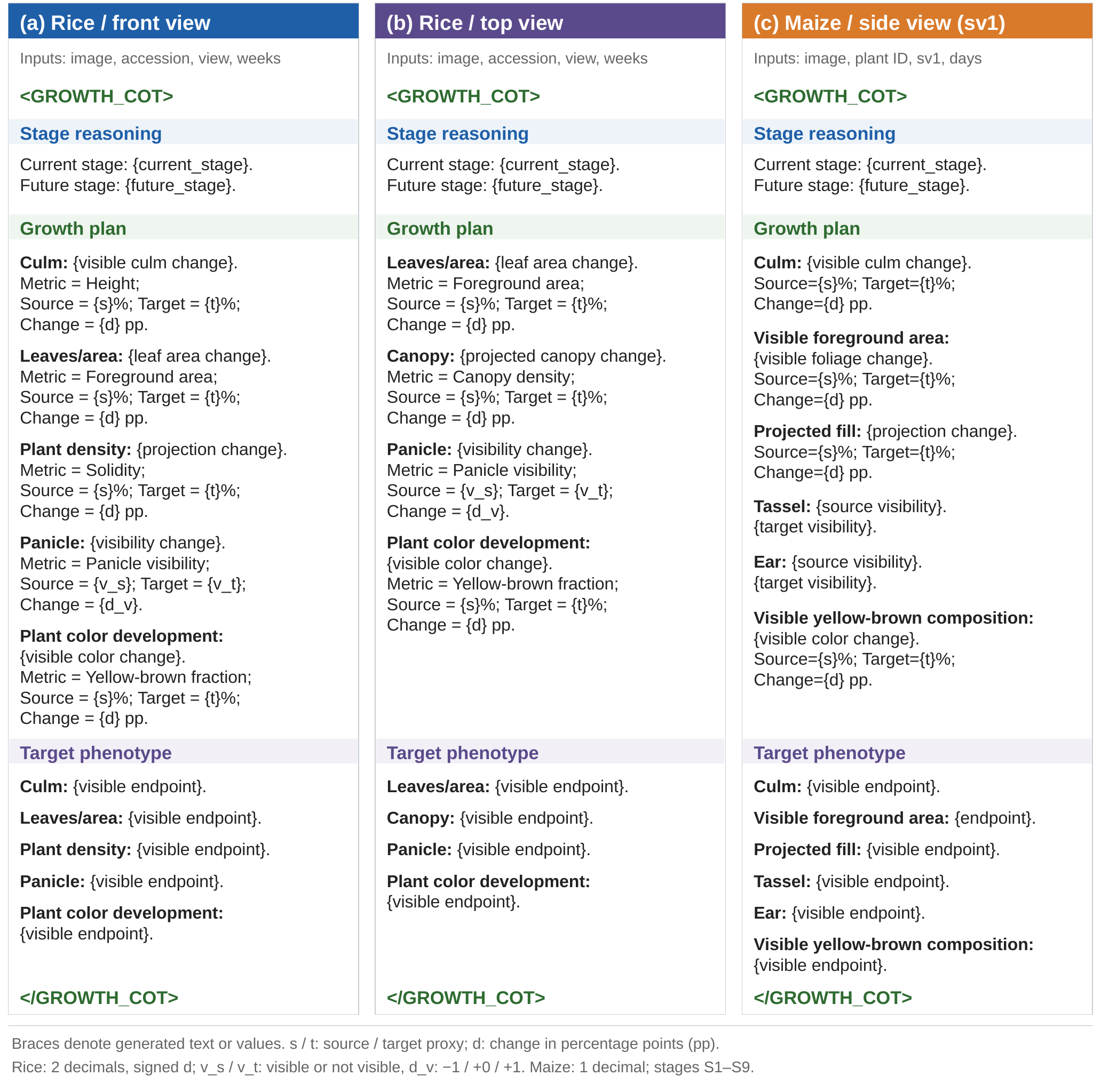}
  \caption{\textbf{CoT output templates for rice and maize.} Each column shows the complete three-section wrapper and ordered fields. Braces denote generated descriptions or values rather than literal output tokens; each field occupies one logical line and wraps here for readability. Growth plan describes the transition and gives proxy evidence, while Target phenotype describes the visible endpoint qualitatively. Image-proxy definitions are in \cref{tab:slots}.}
  \label{fig:supp-cot-templates}
\end{figure}

Each response contains one \texttt{<GROWTH\_COT>} block with three sections in order: \emph{Stage reasoning}, \emph{Growth plan}, and \emph{Target phenotype}.
\Cref{fig:supp-cot-templates} shows the crop- and view-specific templates with their ordered fields and value syntax; current/future stages and numeric targets are predictions, not labels supplied with the query.

For rice, the Planner receives the source image, accession, view, and rounded interval in weeks.
Growth plan fields may include an optional \emph{Guardrails:} clause after the change description; unavailable metric evidence is omitted.
Rice numeric proxies use two decimal places, while Panicle uses \emph{visible}/\emph{not visible} and a signed visibility change.
The top-view template uses Canopy density and omits the front-view Culm/Height field.

\ifmaize
For maize, the query supplies the source image, plant ID, side view, and exact interval in days.
The four numeric fields use one decimal place; Tassel and Ear contain only qualitative visibility descriptions, and hidden organs are not reconstructed.
When the Culm change is within 2\% of the source proxy, its numeric triplet may be replaced by the exact wording \emph{Stable within the registered two-percent proxy tolerance; Source/Target/Change suppressed.}
The maize Planner's system prompt includes a general guide listing all nine stage IDs S1--S9, their corresponding growth stages, and visible descriptions; it does not identify the current or future stage of the query plant.
The dataset protocol maps S1--S9 to V3, V5, V7, V10, V14, VT, R1, R3, and R6 at nominal 25, 32, 50, 59, 70, 78, 85, 91, and 113 days after sowing; these protocol reference days do not imply that every plant was imaged on those dates.
\fi

\FloatBarrier
\section{Evaluation Coverage}
\label{sec:supp-coverage}

The table reports the valid query count for each score in the paper; unavailable phenotype measurements remain missing and are not counted as zero error.
Rice has 2{,}000 validation queries in total. Height applies to front views only, while DINOv2 and stage/event retrieval require an eligible same-plant, same-view photo series.

\begin{table}[!htbp]
  \centering
  \caption{\textbf{Evaluation coverage in the paper.} Counts are valid queries over the indicated dataset.}
  \label{tab:supp-coverage}
  \small
  \setlength{\tabcolsep}{3pt}
  \begin{tabularx}{\linewidth}{@{}p{3.6cm}X@{}}
    \toprule
    Dataset / method & Valid queries by score \\
    \midrule
    Rice, most methods & $N=2{,}000$: area, density, yellow errors; IoU, PSNR, SSIM; $N=1{,}000$: height (front); $N=1{,}964$: DINOv2, Stage R@1, Event R@1 \\
    Rice, FLUX 3 & $N=998$: height; $N=1{,}996$: density and yellow \\
    \ifmaize
    Maize, all methods & $N=1{,}600$: DINOv2, Stage R@1, pixel scores \\
    Maize, copy / GPT Image 2 & $N=1{,}600$: area, culm-proxy height, density, yellow errors \\
    Maize, Simple / AgroWM & $N=1{,}596$ each: area, culm-proxy height, density; $N=1{,}592/1{,}591$: yellow error \\
    \fi
    \bottomrule
  \end{tabularx}
\end{table}

\ifmaize
Maize phenotype counts are lower for trained image models when the foreground is empty or does not contain enough valid color pixels; the underlying predictions remain in the evaluation set.
The maize evaluation uses one view, and its gallery has one image per plant, view, and stage, so Stage and Event retrieval hits coincide; the paper reports Stage R@1.
\fi

\clearpage

\end{document}